\pdfoutput=1
\documentclass[11pt]{article}

\usepackage[final]{acl}

\usepackage{times}
\usepackage{latexsym}

\usepackage[T1]{fontenc}
\usepackage[utf8]{inputenc}

\usepackage{microtype}

\usepackage{inconsolata}

\usepackage{graphicx}

\title{Text2Cypher: Bridging Natural Language and Graph Databases}

\author{Makbule Gulcin Ozsoy \\
  Neo4j / London, UK \\
  \texttt{makbule.ozsoy@neo4j.com} \\\And
  Leila Messallem \\
   Neo4j / Malmö, Sweden \\
  \texttt{leila.messallem@neo4j.com} \\  \AND
  Jon Besga \\
   Neo4j / London, UK \\
  \texttt{jon.besga@neo4j.com} \\\And
  Gianandrea Minneci \\
   Neo4j / London, UK \\
  \texttt{gianandrea.minneci@neo4j.com} 
  \\
}

\begin{document}
\maketitle
\begin{abstract}

Knowledge graphs use nodes, relationships, and properties to represent arbitrarily complex data. When stored in a graph database, the Cypher query language enables efficient modeling and querying of knowledge graphs. However, using Cypher requires specialized knowledge, which can present a challenge for non-expert users. Our work Text2Cypher aims to bridge this gap by translating natural language queries into Cypher query language and extending the utility of knowledge graphs to non-technical expert users.

While large language models (LLMs) can be used for this purpose, they often struggle to capture complex nuances, resulting in incomplete or incorrect outputs. Fine-tuning LLMs on domain-specific datasets has proven to be a more promising approach, but the limited availability of high-quality, publicly available Text2Cypher datasets makes this challenging. In this work, we show how we combined, cleaned and organized several publicly available datasets into a total of 44,387 instances, enabling effective fine-tuning and evaluation. Models fine-tuned on this dataset showed significant performance gains, with improvements in Google-BLEU and Exact Match scores over baseline models, highlighting the importance of high-quality datasets and fine-tuning in improving Text2Cypher performance.

\end{abstract}

\section{Introduction}

\begin{figure}
    \centering
    \includegraphics[width=0.95\linewidth]{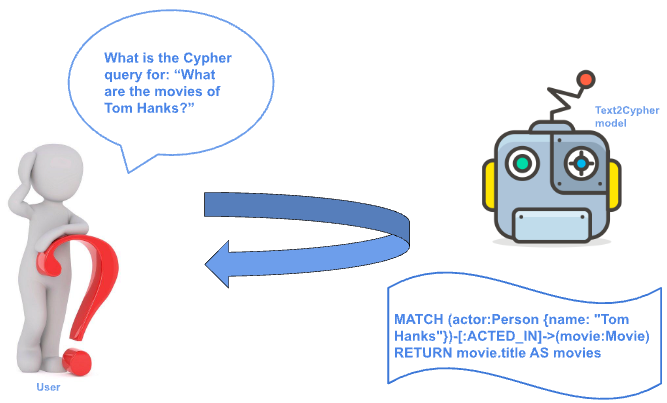}
    \caption{User wants to write a Cypher query for `What are the movies of Tom Hanks`. A Text2Cypher model translates the input natural language question into Cypher, i.e., `MATCH (actor:Person \{name: "Tom Hanks"\})-[:ACTED\_IN]->(movie:Movie) RETURN movie.title AS movies`}
    \label{fig:text2cypher}
\end{figure}

Databases are essential in applications, supporting data storage and knowledge management, and are typically accessed via domain-specific languages like SQL (for relational databases) or Cypher (for graph databases). With advancements in natural language processing, users can now query databases using natural language through applications that perform tasks such as Text2SQL or Text2Cypher. Consequently, even with minimal technical expertise, users can easily retrieve information, build applications such as dashboards or analytics, or integrate knowledge into other systems, like Retrieval-Augmented Generation (RAG).
The Text2Cypher task converts plain language questions into Cypher query language (see Figure \ref{fig:text2cypher}). In the figure, a user wants to write a Cypher query for \textit{`What are the movies of Tom Hanks`}. A Text2Cypher model translates the input natural language question into Cypher, i.e., it returns \textit{`MATCH (actor:Person \{name: "Tom Hanks"\})-[:ACTED\_IN]->(movie:Movie) RETURN movie.title AS movies`}. This generated Cypher query can then be used to retrieve relevant data from the database, allowing for utilization based on the needs of the user.

Foundational large language models (LLMs) can be utilized for Text2Cypher task directly with an appropriate prompt. However, they may struggle with complex, multi-hop queries, leading to incomplete or incorrect outputs which damage the utility of the knowledge graph.
Fine-tuning LLMs on domain-specific datasets offers a promising solution but requires high quality data that pairs natural language queries with Cypher translations, plus schema information for greater accuracy.
However, creating such a dataset is challenging, as it requires an understanding of graph representation, domain-specific knowledge to formulate effective natural language questions, and proficiency in Cypher syntax. \cite{zhong2024synthet2c}. If the training set does not include high-quality, diverse and sufficient examples, the fine-tuned Text2Cypher model may underperform.

The number of publicly available Text2Cypher datasets is limited.  A few examples include those created by Neo4jLabs \cite{neo4jLabsDatasets}, datasets converted from Text2SQL sets \cite{zhao2023rel2graph, zhao2023cyspider, semanticParser4Graph}, and others constructed synthetically \cite{zhong2024synthet2c}. However, these datasets are prepared independently, which makes it difficult to use them together. 
In this work, we combine and refine instances from publicly available datasets, creating a large dataset for training and testing, and use it to benchmark and fine-tune foundational models for Text2Cypher.
Our main contributions are as follows:
\begin{itemize}
    \item We combine instances from publicly available datasets, refining and organizing them to enhance usability. The final dataset includes $44,387$ instances, with a training and test split, of $39,554$ instances for training and $4,833$ for testing. The dataset is made available to the public \footnote{Dataset: \url{https://huggingface.co/datasets/neo4j/text2cypher-2024v1}}.
    \item We use this new dataset to benchmark a range of foundational and previously fine-tuned models on the Text2Cypher task. The results showed that large-foundational models performed the best, however, the fine-tuned models showed promise for improving performance.
    \item We fine-tuned a set of selected foundational models using the new dataset and compared their performance to benchmark results. The results showed that all the fine-tuned models achieve better results than their baseline models. Selected fine-tuned models are made publicly accessible \footnote{A finetuned model: \url{https://huggingface.co/neo4j/text2cypher-gemma-2-9b-it-finetuned-2024v1}}.
\end{itemize}

The paper is structured as follows: Section \ref{rel_work} discusses related work on translating natural language to query languages, with a focus on Text2Cypher. Section \ref{dataset} details the dataset preparation process. Section \ref{benchmark} and Section \ref{finetune} present our experiments for benchmarking and fine-tuning. Finally, Section \ref{conc} concludes the paper.

\section{Related Work}\label{rel_work}
This section provides a brief overview of graph databases and the Cypher query language, along with a summary of natural language to code generation initiatives, and a detailed discussion of efforts related to the Text2Cypher task.

% Check papers listed in refs

% \begin{enumerate}
%     \item Why developers/researchers need Text2Sql, Text2Code etc.
%     \item Works done in Text2Sql, Text2Code etc.
%     \item What is Cypher, definition of graph, graphs database
%     \item Works done in Text2Cypher
% \end{enumerate}

\subsection{Graph Databases and Cypher Language}

\begin{figure}
    \centering
    \includegraphics[width=0.8\linewidth]{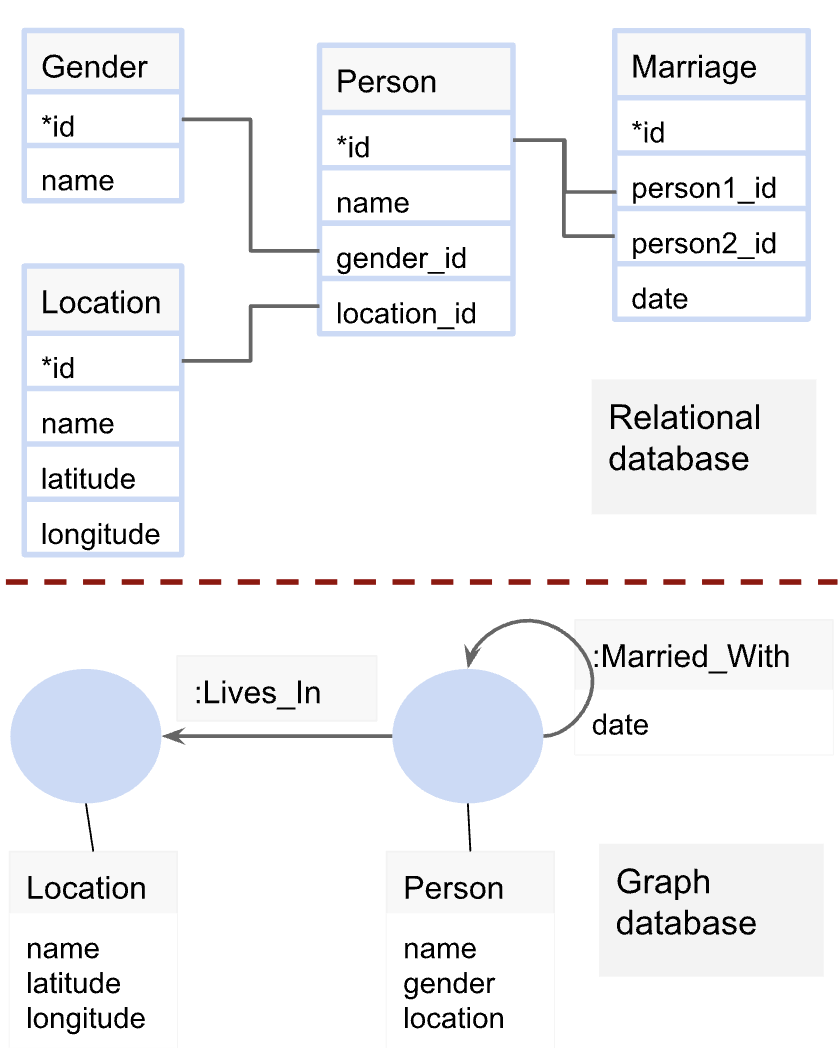}
    \caption{Relational databases uses SQL-based query languages, while Graph databases commonly uses Cypher query language. The figure shows an example representation of Person, Location, Gender and Marriage entities and relationships on a relational and graph database.}
    \label{fig:sql_vs_cypher}
\end{figure}

Graph Database Systems (GDBs) store, manage, and retrieve graph data, where nodes, relationships, and their properties are used for representing real-world knowledge \cite{zheng2024testing}. These systems enable efficient querying of relationships and offer easy visualization \cite{yoon2017use}.

The Figure \ref{fig:sql_vs_cypher} illustrates knowledge representation in graph vs. relational databases, showing people, their locations, and marital relationships. Relational databases require multiple tables and joins to model relationships, while graph databases use nodes (e.g., Person and Location) and edges, simplifying relationship queries. For example, to answer a question like "What is the most common location where married couples live?", a relational database would need several joins, while a graph database can achieve this in a single query. This has lead to graph databases becoming a common underlay for knowledge graphs.

Companies specializing in graph databases include Neo4j \cite{neo4j}, NebulaGraph \cite{wu2022nebula}, and Amazon Neptune \cite{bebee2018amazon}. In April 2024, GQL standard (ISO/IEC 39075:2024) \cite{gql} was released, providing a unified query language for graph databases. The ISO GQL standard is heavily influenced by Neo4j's Cypher language (both share a large amount of syntax and they are both declarative pattern-matching languages). So while this work focuses on translating natural language into Cypher queries, the general approach will be applicable to GQL when it is more widespread.

%This work primarily focuses on Neo4j, one of the most widely used graph databases, along with its query language, Cypher.

\subsection{Natural Language to Code Generation}
Converting natural language to executable code is essential for applications like database interfaces and virtual assistants \cite{pasupat2015compositional, yu2018spider, zhong2024synthet2c, fan2024metasql, agashe2019juice}. Advancements in large language models (LLMs) have enabled significant progress in translating natural language into query languages like SQL or Cypher. This capability allows users with limited technical expertise to to retrieve information, build dashboards, and integrate database knowledge into systems like Retrieval-Augmented Generation (RAG).

There has been extensive research on the Text2SQL task, which translates natural language queries to  SQL queries \cite{yu2018spider, guo2019towards, rajkumar2022evaluating, li2023resdsql, fan2024metasql, li2024can}. In contrast, there is less work focused on the Text2Cypher task, which translates natural language queries into Cypher queries. 
This disparity stems from SQL’s dominance in relational databases and traditionally high industry demand \cite{sqlvscypher}. However, graph-based data representation is not only a more obvious fit for knowledge graphs, but is gaining recognition for addressing issues like hallucinations in RAG models. As such interest in Cypher is increasing, and Cypher’s efficiency in expressing complex, interconnected queries makes it a compelling alternative to SQL for knowledge graphs (and other domains).

% This disparity may be attributed to the differences between these languages \cite{sqlvscypher} (see Figure \ref{fig:sql_vs_cypher}), as SQL queries are predominantly used in relational databases, which have a large community and high industry demand. However, there is a growing recognition of graph structures for data representation, using entities and their relationships, particularly in addressing hallucination issues in Retrieval-Augmented Generation (RAG) models. As a result, Cypher is gaining increased interest and acknowledgment. Additionally, Cypher's ability to efficiently express complex queries about interconnected data makes it a powerful alternative to traditional SQL.

\subsection{Text2Cypher Task}

The Text2Cypher task translates natural language queries into Cypher queries (see Figure \ref{fig:text2cypher}). Large language models (LLMs) can handle this with zero- or few-shot prompts, which have shown promise but are still imperfect \cite{chen2021evaluating}. Fine-tuning LLMs offers a more robust alternative, though it is limited by the scarcity of relevant datasets and high computational costs \cite{ni2023lever}. Some research has focused on creating datasets for Text2Cypher, while others have concentrated on model benchmarking and fine-tuning for this task.

% \paragraph{\textbf{Dataset Preparation}:} 
Some dataset preparation efforts for Text2Cypher involve translating existing datasets from other query languages, while others focus on creating dedicated datasets.
Examples of translations include S2CTrans \cite{zhao2023s2ctrans}, which converts SPARQL queries into Cypher in order to handle complex graph queries, and CySpider \cite{zhao2023cyspider} and Rel2Graph \cite{zhao2023rel2graph}, which map SQL queries to Cypher and create parallel corpora of natural language-to-Cypher pairs.
Specific Text2Cypher datasets include Neo4jLabs datasets \cite{neo4jLabsDatasets}, which are generated via LLMs and their crowd-sourcing tool \cite{neo4jLabsCrowdsourcing}. Opitz et al. \cite{opitz2022zero} and SyntheT2C \cite{zhong2024synthet2c} used synthetic methods to generate Cypher query data. 
While several efforts have been made to create datasets for the Text2Cypher task, these datasets are often developed independently. In this work, we aim to compile a well-structured Text2Cypher dataset by combining and structuring instances from publicly available sources.

% \begin{table*}
%   \caption{Input data sources}
%   \label{tab:data_source}
%   \begin{tabular}{lll}
%     \hline
%     \textbf{Name} &\textbf{Alias} & \textbf{SourceType} \\
%     \hline
%     Functional Cypher Generator & neo4jLabs\_text2cypher\_functionalCypher & Neo4jLabs \cite{neo4jLabsDatasets}\\
%     Synthetic gemini demodbs & neo4jLabs\_text2cypher\_gemini & Neo4jLabs \cite{neo4jLabsDatasets}\\
%     Synthetic gpt4o demodbs & neo4jLabs\_text2cypher\_gpt4o & Neo4jLabs \cite{neo4jLabsDatasets}\\
%     Synthetic gpt4turbo demodbs  & neo4jLabs\_text2cypher\_gpt4turbo & Neo4jLabs \cite{neo4jLabsDatasets}\\
%     Synthetic opus demodbs & neo4jLabs\_text2cypher\_claudeopus & Neo4jLabs \cite{neo4jLabsDatasets}\\
%     Rag-Eval datasets & neo4j\_rageval\_[movies,products]\_text2cypher\_results & Neo4j \\
%     Neo4j-Text2Cypher’23 datasets & neo4j\_text2cypher2023-[train, test] & Neo4j \\
%     Crowdsourcing dataset & neo4j\_crowdsourced\_text2cypher\_raw & Neo4j \\
%     HF-iprahara/text\_to\_cypher & hf\_iprahara\_text\_to\_cypher & HuggingFace (HF)\\
%     HF-dfwlab/cypher & hf\_dfwlab\_cypher\_eng-to-cypher & HuggingFace (HF)\\
%     HF-vedana17/text-to-cypher & hf\_vedana17\_text-to-cypher\_dataset & HuggingFace (HF)\\
%     Cy-Spider & cySpider\_semanticParser4Graph\_data\_folder & SemanticParser4Graph \cite{semanticParser4Graph} \\

%   \hline
% \end{tabular}
% \end{table*}

% \paragraph{\textbf{Benchmarking and Fine-tuning models}} 
Some research has focused on benchmarking and fine-tuning models for the Text2Cypher task:
Neo4jLabs \cite{neo4jLabsModels} released fine-tuned models based on their datasets, using LLMs like LLama and Codestral. 
GPT4Graph \cite{guo2023gpt4graph} evaluated LLMs on graph tasks, including Cypher query generation, using the MetaQA \cite{zhang2017variational} dataset and testing InstructGPT-3 \cite{ouyang2022training} in zero- and one-shot settings. 
TopoChat \cite{xu2024topochat} developed a material sciences dataset, using prompts to generate Cypher queries with foundational LLMs.
Baraki et al. \cite{baraki2024leveraging} leveraged Neo4jLabs' crowd-sourced and synthetic datasets to fine-tune models, using the crowd-sourced set for evaluation. 
TransKGQA \cite{chong2024transkgqa} extracted information from knowledge graphs, using the ‘sentence-transformers/all-MiniLM-L12-v2’ model to generate Cypher queries. 
Although these works have provided fine-tuned models, the number of models used was limited. In our work, after constructing a larger and more organized dataset, we benchmark and fine-tune a wider range of baseline LLMs.

% In this work, we aim to create a well-structured Text2Cypher dataset by combining instances from publicly available datasets. We will utilize this dataset for benchmarking and fine-tuning new models. The subsequent sections will outline our data preparation process and present our experimental results.

\section{Dataset Construction}\label{dataset}
While several Text2Cypher datasets exist, many are prepared separately, making them hard to use together. In this work we bring instances from publicly available datasets together, clean and organize them for smoother use. For this purpose, we executed three main steps: (i) Identification and collection of publicly available datasets, (ii) Combining and cleaning the data, and (iii) Creating the training and test splits.

\subsection{Identification and collection of publicly available datasets}
As the initial step, we  identified the datasets which are already publicly available. We have identified 25 different resources from (i) Neo4j resources (including Neo4jLabs) (ii) HuggingFace (HF) datasets and (iii) Academic papers. Out of these resources, we were able to utilize 16 of those datasets, as they met our criteria of including natural language question and Cypher query pairs, as well as database schema information, along with appropriate licensing and accessibility. % The list of data sources are presented in Table \ref{tab:data_source}.

\begin{table}
  \caption{Data fields}
  \label{tab:fields}
  \begin{tabular}{p{0.35\linewidth}p{0.55\linewidth}}
    \hline
    \textbf{Field name} &\textbf{Description} \\
    \hline
    question & Textual question \\ %E.g., “What is the total number of companies?” \\
    schema & The database schema \\
    cypher & Output cypher query \\ %E.g., “MATCH (c:Company) RETURN COUNT(c)” \\
    data\_source & Alias of the dataset source \\ % E.g., "neo4jLabs\_synthetic\_gpt4turbo"
    database\_reference  & Alias of the database \\ %  \_alias (if available) E.g., None, "neo4jlabs\_demo\_db\_movies" 
    instance\_id & Incremental index  \\ % assigned per row
  \hline
\end{tabular}
\end{table}

\subsection{Combining and cleaning the data}
After identifying the input datasets, we standardized them into a single format. Each row was reformatted to include fields ["question", "schema", "cypher", "data\_source", "database\_reference", "instance\_id"], as described in Table \ref{tab:fields}. 
One of the fields, namely "database\_reference", requires particular attention. In some cases within the combined dataset, database access is available where the reference or the generated Cypher queries can be executed. Further details about these databases can be found at the page of Neo4jLabs-Crowdsourcing Initiative \cite{neo4jLabsCrowdsourcing}.

The combined dataset is further cleaned in two steps:
\begin{itemize}
    \item \textbf{Manual checks and updates}: This step aims to produce more reliable and error-free output data. Queries are manually reviewed, and errors are corrected through straightforward removals or updates: (i) Updating Cypher queries, such as removing unwanted characters (e.g., back-tick) (ii) Removing irrelevant questions (e.g., "Lorem ipsum …") (iii) Deduplicating rows based on the ["question", "cypher"] pairs.
    \item \textbf{Syntax validation}: Each Cypher query is checked for syntax errors by running 'EXPLAIN` clauses in a local Neo4j database. Queries that trigger syntax errors are identified and removed from the combined dataset. Additionally, the queries are de-duplicated.
\end{itemize}

\begin{table*}
  \caption{Models used for benchmarking}
  \label{tab:benchmark}
  \begin{tabular}{llll}
    \hline
    \textbf{Type} &\textbf{Name} & \textbf{Base model}  \\
    \hline
    HF & hf\_ft\_lakkeo\_stable\_cypher\_instruct3B & Stability AI/Stable-code-instruct-3b\\
    HF & hf\_ft\_tomasonjo\_text2cypher & Meta/Llama-3-8b-Instruct\\
    HF & hf\_ft\_neo4j\_text2cypher\_23\_codellama & Meta/CodeLlama\-13B \\
    OpenAI & openai\_ft\_neo4j\_text2cypher\_23\_gpt3\_5 & OpenAI/GPT\-3.5 \\
    \hline
    HF & hf\_foundational\_meta\_llama3\_1\_8B\_instruct & Meta/LLama-3.1-8B-instruct  \\
    HF & hf\_foundational\_codeLlama\_7B\_instruct\_hf & Meta/CodeLLama-7B-instruct  \\
    HF & hf\_foundational\_gemma2\_9B\_it & Google/Gemma-2-9B-it  \\
    HF & hf\_foundational\_codegemma\_7B\_it & Google/CodeGemma-7B-it  \\
    \hline
    OpenAI & openai\_gpt3\_5 & OpenAI/GPT-3.5  \\
    OpenAI & openai\_gpt4\_o & OpenAI/GPT-4o  \\
    OpenAI & openai\_gpt4\_o\_mini & OpenAI/GPT-4o-mini  \\
    VertexAI & gemini-1.0-pro-002 & Google/Gemini-1.0-Pro  \\
    GoogleAIStudio & gemini-1.5-flash-001 & Google/Gemini-1.5-Flash  \\
    GoogleAIStudio &  gemini-1.5-pro-001 & Google/Gemini-1.5-Pro  \\
  \hline
\end{tabular}
\end{table*}

\begin{figure}
    \centering
    \includegraphics[width=0.99\linewidth]{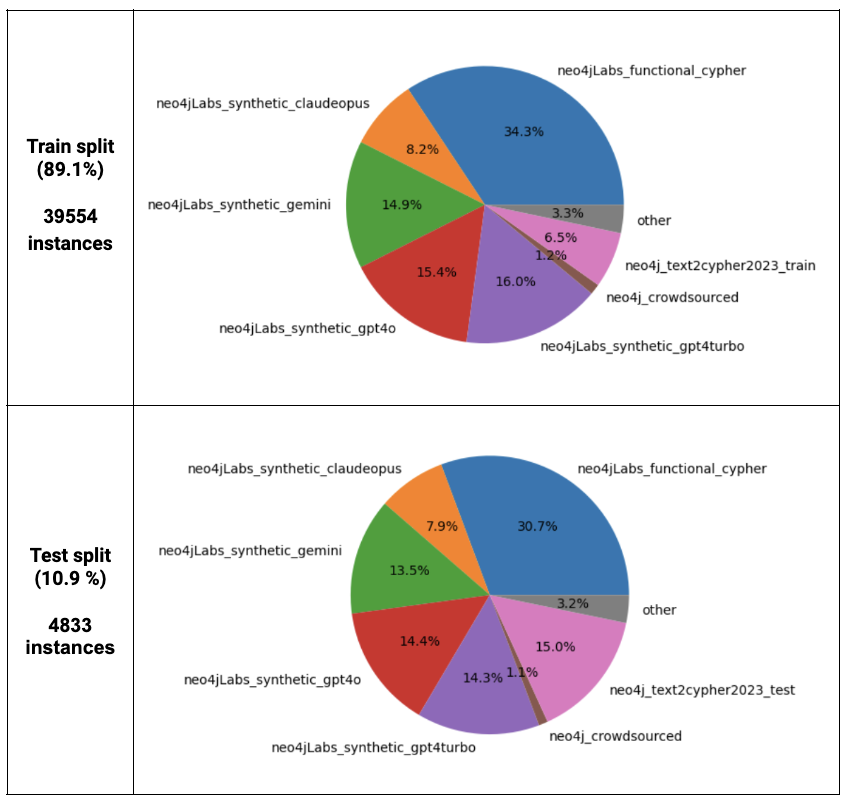}
    \caption{Data distribution: The train and test splits consist ${\sim}89\%$ and ${\sim}11\%$ of the overall data, respectively.}
    \label{fig:data_distribution}
\end{figure}

\subsection{Creating the training and test splits}
Having the cleaned dataset, the final step is to prepare the training and test splits. We have identified 3 groups of datasets:
(i) Train-specific datasets: Files with "train" in the name, used for training. 
(ii) Test-specific datasets: Files with "test" or "dev" in the name, used for testing. 
(iii) Remaining datasets: Files with no specified use. 
We assigned Train-specific datasets to the training split and Test-specific datasets to the test split. The remaining datasets were split 90:10 for training and testing, respectively. Each split was shuffled to prevent over-fitting from sequence or repetitive questions. 
% Note that even though a specific validation set is not provided, the researcher are encouraged to use a validation split during their work.

The data preparation resulted in $44,387$ instances, with $39,554$ instances in the training split and $4,833$ instances in the test split. The train and test splits consist ${\sim}89\%$ and ${\sim}11\%$ of the overall data, respectively. Their distribution across data sources is similar, as shown in Figure \ref{fig:data_distribution}. 
As explained previously, not every instance in the training and test sets has database access, as indicated by the "database\_reference" field. Analyzing the distribution of instances with database access reveals that the training set contains $22,093$ such instances ($55.85\%$ of the total), while the test set has $2,471$ instances ($51.12\%$ of the total). These instances are later used in the experimentation with an additional evaluation procedure.

% TODO May add: ---> Note: Even though deduplication is done on [question, cypher] pairs, it was possible to have same ‘question’s in training and test set. The updated version of test split removed such instances; test split (with removed seen questions): 3306 instances.

\section{Model Evaluation and Benchmarking} \label{benchmark}
After constructing a larger and more organized dataset, this section presents the benchmarking results.
% this section outlines the experimental setup and presents the benchmarking results. %a wider range of baseline LLMs.

% In the following sections, we will provide detailed information about the experimental setup and the evaluation results.

\subsection{Evaluation metrics} \label{eval_metrics}

Text2Cypher is a type of text-to-text generation task, where natural language questions are translated into Cypher queries. Therefore, evaluation metrics commonly used in other text-to-text tasks, such as machine translation and summarization, can also be applied to this task. 
Using HuggingFace Evaluate library \cite{hfEvaluate}, we computed: (i) Text2Text comparison metrics, such as ROUGE, BLEU, METEOR (ii) Embedding similarity metrics, such as BERTScore, FrugalScore (iii) Text similarity metrics, such as Cosine and Jaro-Winkler similarity, and (iv) Exact Match score.  Although we calculated all these metrics, we primarily use Google-BLEU and Exact Match scores throughout the paper.
%In the future, we plan to explore the HuggingFace LightEval library \cite{lighteval}, which is recommended for evaluating LLMs.

\subsection{Experimental Setup}
For benchmarking the models, we used the test split of the larger dataset introduced in Section \ref{dataset}. Closed models were evaluated through APIs provided by the respective companies. For the other models, which are openly accessible via HuggingFace (HF), we utilized HF interfaces. To access GPUs for evaluation, we employed RunPod \cite{runpod} environments. Where relevant, we followed the instructions outlined in Table \ref{tab:instructions}, which were inspired from tips provided by Neo4jLabs \cite{neo4jLabsLangChainTips}.

\begin{table}
  % \caption{Instructions used, which were inspired from tips provided by Neo4jLabs \cite{neo4jLabsLangChainTips}}
\caption{Instructions used}
  \label{tab:instructions}
  \begin{tabular}{p{0.15\linewidth}p{0.78\linewidth}}
    \hline
    \textbf{Type} & \textbf{Instruction prompt}  \\
    \hline
    System \newline Instruct. &  Task: Generate Cypher statement to query a graph database. Instructions: Use only the provided relationship types and properties in the schema. Do not use any other relationship types or properties that are not provided in the schema. Do not include any explanations or apologies in your responses. Do not respond to any questions that might ask anything else than for you to construct a Cypher statement. Do not include any text except the generated Cypher statement. \\
    \hline
    User \newline Instruct. & Generate Cypher statement to query a graph database. Use only the provided relationship types and properties in the schema. \newline
            Schema: \{schema\} \newline
            Question: \{question\} \newline
            Cypher output: 
         \\
  \hline
\end{tabular}
\end{table}

We defined two types of evaluation procedures: 
\begin{itemize}
    \item \textbf{Translation-based evaluation}: The generated Cypher queries are compared with the reference Cypher queries based solely on the textual content. The evaluation metrics used for this comparison are detailed in Section \ref{eval_metrics}. 
    \item \textbf{Execution-based evaluation}: The generated and reference Cypher queries are executed on the target databases, and their outputs are collected. Afterward, the execution results are converted into string representations (ordered lexicographically for consistency). The same evaluation metrics used in the translation-based evaluation are then applied to these outputs.
\end{itemize}

% \subsection{Models for Comparison} 
% For benchmarking, we aimed to evaluate not only baseline LLMs but also previously fine-tuned models specifically tailored for the Text2Cypher task. To achieve this, we explored models from HuggingFace (HF), academic papers, Neo4jLabs, and various internal Neo4j projects. We identified four fine-tuned models and ten foundational models (e.g., GPT-4o, CodeLlama) that emerged as strong candidates for comparison and benchmarking. The list of models used for benchmarking purpose are listed in Table \ref{tab:benchmark}. In the table, first group includes the fine-tuned models, second group includes the open-weighted models and the last group includes the closed models. 
% We aimed to maintain similar model sizes whenever possible, particularly for the open-weighted models. In the future, we plan to explore both larger and smaller models to compare their performance. By assessing a wider range of model sizes, we hope to gain insights into how size impacts performance in the Text2Cypher task.

\subsection{Benchmarking results}
For benchmarking, we aimed to evaluate not only baseline LLMs but also previously fine-tuned models specifically tailored for the Text2Cypher task. The list of models used for benchmarking purpose are listed in Table \ref{tab:benchmark}.  In the table, first group includes the fine-tuned models, second group includes the open-weighted models and the last group includes the closed models. 
%We present Google-BLEU \cite{wu2016googles} score for translation-based evaluation and the Exact Match score for execution-based evaluation. 

\begin{figure*}
    \centering
    \includegraphics[width=0.80\linewidth]{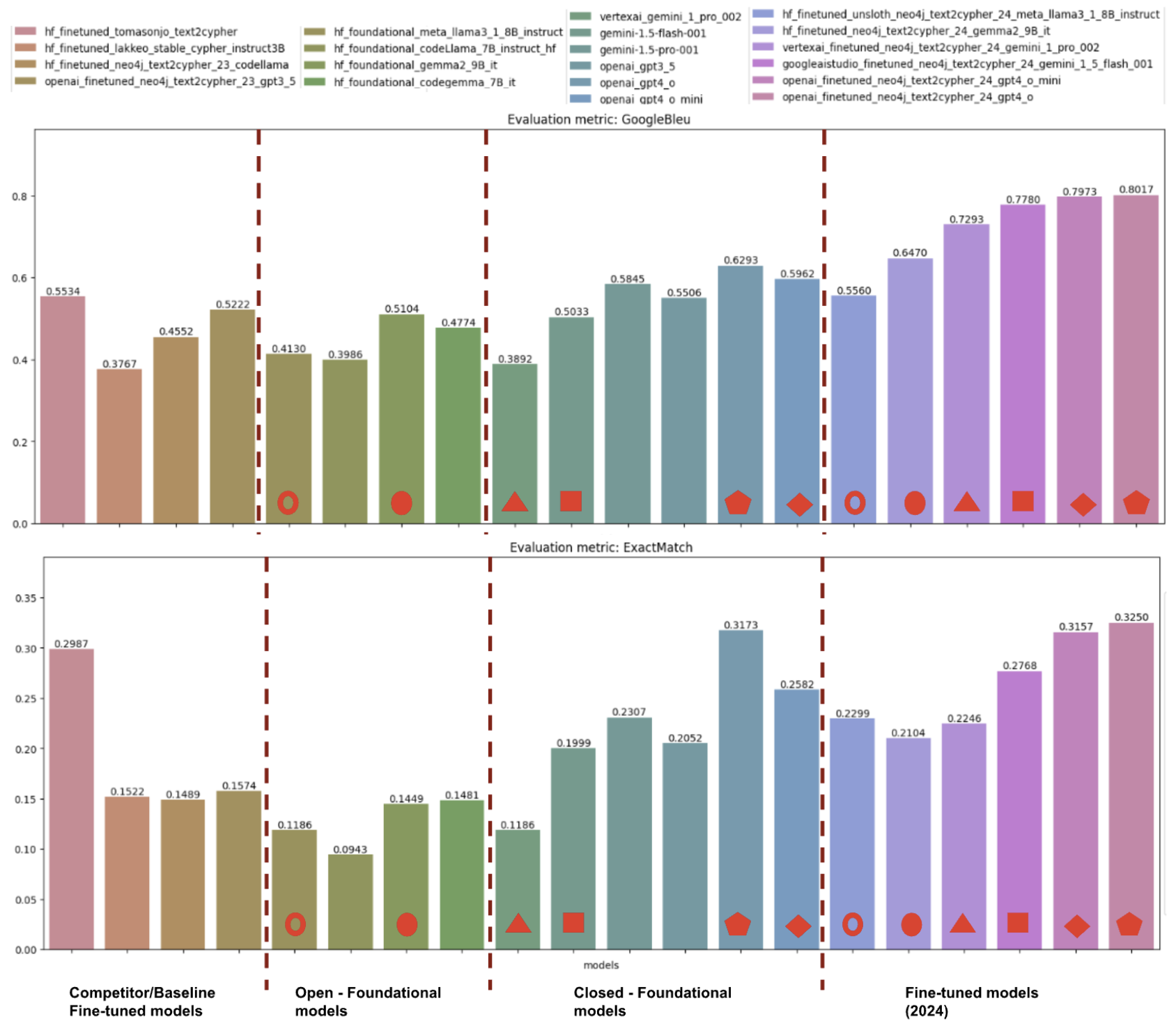}
    \caption{Performance comparison of the baseline and finetuned models}
    \label{fig:finetune_eval}
\end{figure*}

Figure \ref{fig:finetune_eval} presents the performance comparison of the selected models on the test split. We present Google-BLEU score for translation-based and Exact Match score for execution-based evaluation. 
%The models are grouped similarly to Table\ref{tab:benchmark}, with the first group representing fine-tuned models, the second group representing the open-weighted models, and the final group representing the closed models. 
Among the previously fine-tuned models, i.e., with different data, HF/tomasonjo\_text2cypher performed best, but this may be misleading as it had encountered 14.4\% of the test data during training. 
Among the open-weighted models, Google/Gemma-2-9B-it is the best performing model. Contrary to expectations, the code-focused models (e.g., CodeGemma) did not outperform the baseline models. This may be attributed to the fact that Cypher queries are relatively closer to natural language, reducing the advantage of code-specific models.
Among closed-foundational models, the best performing models are OpenAI/GPT4o, OpenAI/GPT4o-mini, and Google/Gemini-1.5-Pro-001 led in performance, with larger models outperforming smaller ones.

Overall, closed foundational models like GPT and Gemini achieved the best performance, though at higher costs. Fine-tuned models improved baseline open-weighted models. In the next section, we explore the process of fine-tuning models and evaluating them using the new dataset introduced in Section \ref{dataset}.

\section{Model Finetuning and Evaluation} \label{finetune}
Based on the findings of benchmarking, presented in Section \ref{benchmark}, we selected six baseline models for our subsequent steps, presented in Table \ref{tab:finetune_models}. In the table, first group includes the open-weighted models, while the second group includes the closed models. Although some models, such as Google/Gemini-1.5-Pro, demonstrated better performance in the benchmark results, they were unavailable for fine-tuning at the time of this analysis and are therefore not included in this work.

\begin{table}
  \caption{Models used for fine-tuning}
  \label{tab:finetune_models}
  \begin{tabular}{ll}
    \hline
    \textbf{Type} & \textbf{Base model}  \\
    \hline
    HF & Meta/LLama-3.1-8B-instruct  \\
    HF &  Google/Gemma-2-9B-it  \\
    \hline
    OpenAI &  OpenAI/GPT-4o  \\
    OpenAI &  OpenAI/GPT-4o-mini  \\
    VertexAI & Google/Gemini-1.0-Pro  \\
    GoogleAIStudio &  Google/Gemini-1.5-Flash  \\
  \hline
\end{tabular}
\end{table}

\subsection{Experimental setup}
For the finetuning process, we used the training split of the larger dataset introduced in Section \ref{dataset}. The closed models were trained through APIs provided by their respective companies, while the other models were finetuned using HuggingFace (HF) or Unsloth \cite{unsloth} interfaces on GPU machines hosted in RunPod \cite{runpod} environments. % It is important to note that this work is part of an ongoing research and exploration, aimed at showcasing the dataset's potential rather than presenting a production-ready solution.
The evaluation procedures and metrics were identical to those used in benchmarking section, Section \ref{benchmark}. The instructions remained consistent with those outlined in Table \ref{tab:instructions}. We used Google-BLEU score for translation-based and Exact Match score for execution-based evaluation.

\subsection{Finetuning results}
The evaluation results for all models, including the previously benchmarked ones, are shown in Figure \ref{fig:finetune_eval}. The last group in the figure highlights the fine-tuned models trained on the dataset introduced in Section \ref{dataset}.  For easier comparison, red shapes are used to link each fine-tuned model with its corresponding baseline version. The figure shows that all the fine-tuned models achieve better results than their baseline models. The best results are obtained by the Finetuned-OpenAI/Gpt4o, Finetuned-OpenAI/Gpt4o-mini and Finetuned-GoogleAIStudio/Gemini-1.5-Flash-001 models.

\begin{table}
  \caption{The improvements of the fine-tuned models over the baseline models}
  \label{tab:finetune_model_improvements}
  \begin{tabular}{p{0.45\linewidth}p{0.20\linewidth}p{0.20\linewidth}}
    \hline
    \textbf{Baseline model} &\textbf{$\Delta$Google BLEU} & \textbf{$\Delta$Exact Match}  \\
    \hline
    HF/LLama3.1-8B-it & ${\sim}0.14$ & ${\sim}0.11$ \\
    HF/Gemma2-9B-it & ${\sim}0.13$ & ${\sim}0.07$ \\
    \hline
    VertexAI/Gemini-1.0-Pro-002 & ${\sim}0.34$ & ${\sim}0.11$\\
    GoogleAIStudio/Gemini-1.5-Flash-001 & ${\sim}0.27$ & ${\sim}0.09$\\
    OpenAI/Gpt-4o-mini  & ${\sim}0.20$ & ${\sim}0.06$ \\
    OpenAI/Gpt-4o & ${\sim}0.18$ & ${\sim}0.01$ \\
    
  \hline
\end{tabular}
\end{table}

The improvements of the fine-tuned models over the baseline models are presented in Table \ref{tab:finetune_model_improvements}. The enhancements for models that already performed well are relatively smaller than for others. For example, OpenAI/Gpt-4 shows an 0.18 increase in the Google-BLEU score, while the VertexAI/Gemini-1.0-Pro-002  demonstrates a 0.34 increase. 
The improvements of the finetuned open-weighted models, i.e. HF/LLama3.1-8B-it and HF/Gemma2-9B-it, are relatively less pronounced. During fine-tuning of these models, our goal was to minimize resource usage (e.g., cost and memory). As a result, with better-tuned parameters, we could potentially achieve even stronger results.

While all the fine-tuned models showed improvements in Google-BLEU and Exact Match scores, it is important to remain aware of the potential risks and pitfalls associated with fine-tuning.

\section{Risks and Pitfalls}
The previous sections demonstrated how fine-tuned models significantly boost performance. However, there are several risks and pitfalls that must be considered.

Even though we de-duplicated the dataset by ["question", "cypher"] pairs, it is still possible to have instances where the same "question" appears with different "cypher" outputs. In such cases, these instances may have been split between the training and test sets, meaning that fine-tuned models could have already encountered the same "question" during training. However, since these instances have different "cypher" outputs, even if the fine-tuned models memorize the "cypher" output for the question, their generated response would be incorrect. This essentially penalizes the models for having seen and memorized the question. In the future, we plan to clean the test set of such instances, re-run the evaluation, and assess any performance differences.

Our dataset is constructed by collecting and combining publicly available datasets, which may include paraphrased versions of the same questions. It is known that training on paraphrased examples of the test set may artificially inflate the performance of the fine-tuned model \cite{yang2023rethinking}. Additionally, both the training and test sets are drawn from the same data distribution, sampled from a larger dataset. If the data distribution shifts, the results may not hold up in the same way.

Finally, the dataset used was gathered from publicly available sources. Over time, foundational models may gain access to both the training and test sets, potentially achieving similar or even better performance results in the future.

% \begin{itemize}
%      \item Even though we deduplicated the dataset by ["question", "cypher"] pairs, it is still possible to have instances where the same "question" appears with different "cypher" outputs. In such cases, these instances may have been split between the training and test sets, meaning that fine-tuned models could have already encountered the same "question" during training. However, since these instances have different "cypher" outputs, even if the fine-tuned models memorize the "cypher" output for the question, their generated response would be incorrect. This essentially penalizes the models for having seen and memorized the question. In the future, we plan to clean the test set of such instances, re-run the evaluation, and assess any performance differences.
%     \item Our dataset is constructed by collecting and combining publicly available datasets, which may include paraphrased versions of the same questions. It is known that training on paraphrased examples of the test set may artificially inflate the performance of the finetuned model \cite{yang2023rethinking}.
%     \item In our evaluation setup, both the training and test sets are drawn from the same data distribution, sampled from a larger dataset. If the data distribution shifts, the results may not hold up in the same way.
%     \item The datasets used were gathered from publicly available sources. Over time, foundational models may gain access to both the training and test sets, potentially achieving similar or even better performance results in the future.
% \end{itemize}

\section{Conclusion} \label{conc}
Databases are essential for data storage, management, and retrieval, typically accessed through domain-specific languages like SQL or Cypher. Recent advancements have made it possible to access databases using natural language through tasks like Text2Cypher. 
While large language models (LLMs) can be applied directly to this task, they often struggle with understanding the nuances of queries, resulting in incomplete or incorrect Cypher outputs. Fine-tuning LLMs on specific Text2Cypher datasets offers a more effective solution. However, publicly available Text2Cypher datasets are limited and often created independently, making them difficult to combine and use effectively.
To address this, we combined and refined available datasets into a unified set of 44,387 instances, with 89\% in the training split and 11\% in testing. 
Using this dataset, we fine-tuned various models, and compared their performance against foundational and previously fine-tuned models, with different datasets. The results revealed that fine-tuned models consistently outperformed their baseline versions, achieving up to a 0.34 increase in Google-BLEU score and a 0.11 increase in Exact Match score. This work highlights the importance of dataset and fine-tuning for Text2Cypher task. Future work will refine this dataset further, analyze challenging cases, and explore model size and prompt engineering effects.

% \section{Limitations}
% % \textcolor{red}{TBD - Maybe?}
% % \textcolor{red}{Check \cite{li2024multisql}}

\bibliography{refs}

% \appendix

% \section{Example Appendix}
% \label{sec:appendix}

% This is an appendix.

\end{document}